\documentclass{article}

\ifdefined\EPAnonymous
  \usepackage[main]{neurips_2026}
\else
  \usepackage[preprint]{neurips_2026}
\fi

\usepackage[utf8]{inputenc}
\usepackage[T1]{fontenc}
\usepackage{textcomp}
\usepackage{hyperref}
\usepackage{url}
\usepackage{booktabs}
\usepackage{amsfonts}
\usepackage{amsmath}
\usepackage{amssymb}
\usepackage{nicefrac}
\usepackage{microtype}
\usepackage{xcolor}
\usepackage{graphicx}
\usepackage{flafter}
\usepackage{algorithm}
\usepackage{algorithmic}
\usepackage{multirow}
\usepackage{subcaption}
\usepackage{tikz}
\usetikzlibrary{positioning, arrows.meta, calc, shapes.geometric}
\usepackage{cleveref}

\usepackage{iftex}
\ifluatex
  \usepackage{fontspec}
\IfFontExistsTF{Noto Sans Arabic} {\newfontfamily{\arabicfont}{Noto Sans Arabic}} {\IfFontExistsTF{Geeza Pro} {\newfontfamily{\arabicfont}{Geeza Pro}} {\IfFontExistsTF{Amiri} {\newfontfamily{\arabicfont}{Amiri}} {\newfontfamily{\arabicfont}{Latin Modern Roman}}}} \IfFontExistsTF{Noto Sans CJK SC} {\newfontfamily{\cjkfont}{Noto Sans CJK SC}} {\IfFontExistsTF{PingFang SC} {\newfontfamily{\cjkfont}{PingFang SC}} {\IfFontExistsTF{Noto Serif CJK SC} {\newfontfamily{\cjkfont}{Noto Serif CJK SC}} {\newfontfamily{\cjkfont}{Latin Modern Roman}}}}
  \usepackage{newunicodechar}
\directlua{ for i = 0x0600, 0x06FF do tex.print("\\newunicodechar{" .. utf8.char(i) .. "}{{\\arabicfont " .. utf8.char(i) .. "}}") end for i = 0x4E00, 0x9FFF do tex.print("\\newunicodechar{" .. utf8.char(i) .. "}{{\\cjkfont " .. utf8.char(i) .. "}}") end } \fi

\title{Exemplar Partitioning for Mechanistic Interpretability}

\author{%
  Jessica Rumbelow \\
  Leap Laboratories \\
  \texttt{jessica@leap-labs.com} \\
}

\begin{document}

\maketitle

\begin{abstract}
We introduce Exemplar Partitioning (EP), an unsupervised method for building interpretable feature dictionaries from large language model activations. An EP dictionary is a Voronoi partition of activation space, built by leader-clustering streamed activations within a distance threshold. Each region is defined by an observed exemplar and an average of its member activations, which define region membership and provide directions for intervention. Dictionary size is determined by the activation stream at the chosen threshold rather than pre-selected. Exemplars link regions to observed inputs, allowing dictionaries built from the same input stream to be compared across layers, training checkpoints, and architectures.

We demonstrate how EP can be used to interpret and intervene on model behaviour, track changes in activation space through training, and detect hidden concepts. Comparing EP dictionaries on base and instruction-tuned Gemma-2-2B and Llama-3.1-8B reveals that instruction tuning reorganises harmful prompt activations similarly across the two models, but at different granularities. Interventions on these regions make both models answer harmful requests they previously refused. In 19 of 21 Taboo models trained to hide a secret word, EP finds new regions that do not exist in the base model, whose decoded tokens relate to the known secret on inspection.

Although EP assigns each token to a single region, linear probes built from EP regions achieve up to $90.5\%$ of full-activation probe accuracy.

On AxBench concept detection at Gemma-2-2B-it layer 20, EP achieves the highest mean AUROC of all unsupervised methods ($0.937$), outperforming SAE-A ($0.911$) and approaching supervised probes ($0.946$). Building EP dictionaries is fast and cheap: EP uses $\sim 10^3$ times fewer construction tokens than comparable SAEs. 

\ifdefined\EPAnonymous
We release all code.
\else
We release all code at \url{https://github.com/jessicarumbelow/exemplar-partitioning}.
\fi
\end{abstract}

\section{Introduction}

\begin{figure}[t]
  \centering
  \includegraphics[width=0.85\linewidth]{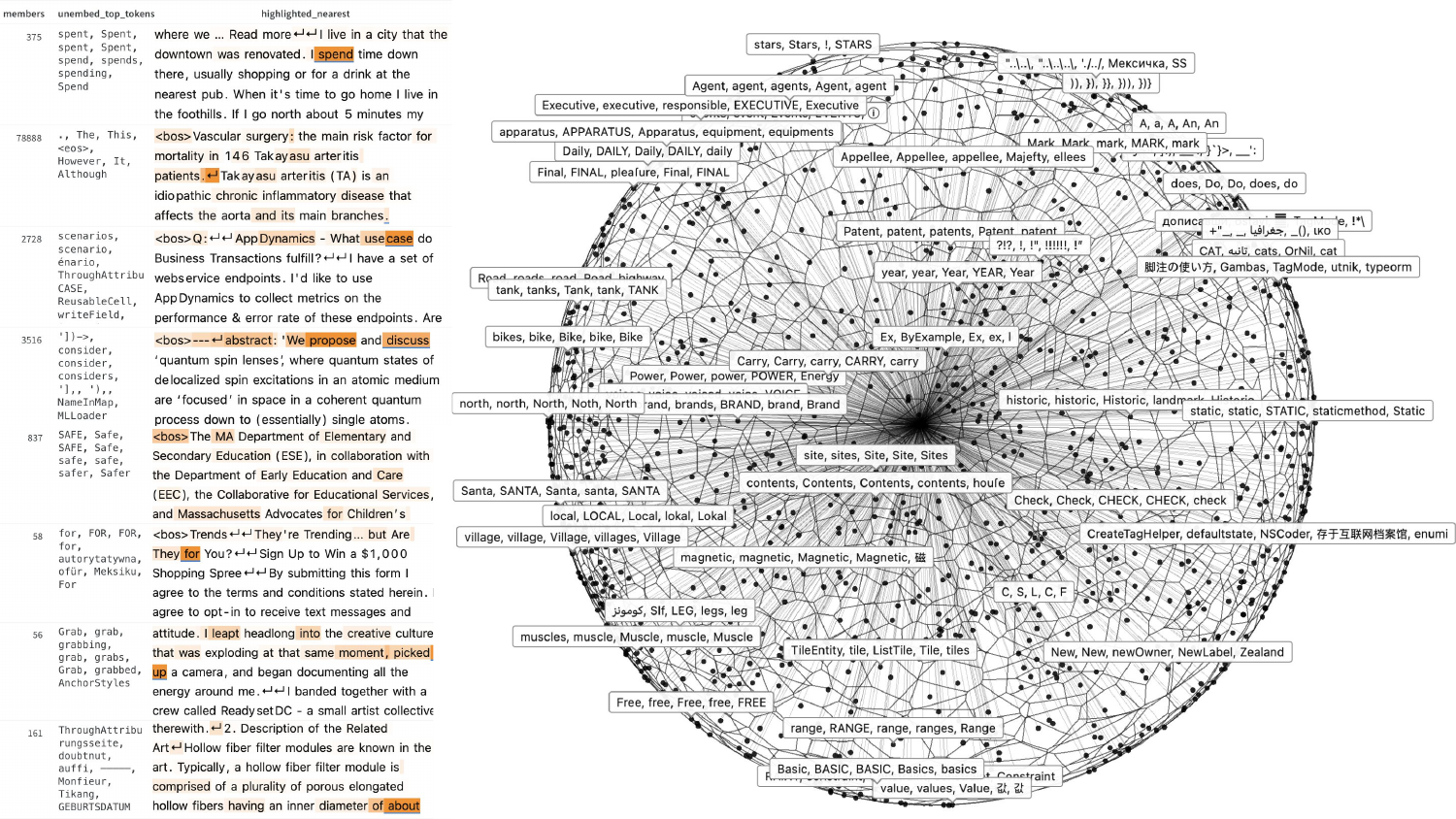}
  \caption{An Exemplar Partitioning dictionary built from Gemma-2-2B L12 activations at $p_{2}$ ($K=5{,}129$). \emph{Left:} eight sample regions, each shown with its member count, its exemplar's logit-lens~\citep{nostalgebraist2020logitlens} decode, and an excerpt of a member input with the activating tokens highlighted. \emph{Right:} a PCA-projected 3D rendering of the Voronoi partition; each cell is one region, with a random selection also labelled with logit-lens decode.}
  \label{fig:headline}
\end{figure}

Unsupervised feature discovery aims to find recurring structure in a model's activations without specifying in advance what to look for, which has important applications in AI safety and scientific discovery \citep{cunningham2023sparse,bricken2023monosemanticity,templeton2024scaling}. The standard tool for this is the sparse autoencoder (SAE), which learns a dictionary of $n$ directions by training an encoder and decoder to reconstruct each activation as a sparse sum of those directions \citep{templeton2024scaling,lieberum2024gemmascope}. This objective commits to a particular understanding of activation space, in which every activation is a sum of a few features, each feature is a direction, and there are $n$ of them. These commitments are natural for decomposition and reconstruction, but not obviously required for interpretability. How much of the utility SAEs provide is accessible by a more direct construction?

We answer this with Exemplar Partitioning (EP), an extremely cheap and simple method which uses leader clustering to create a map of activation space at a given resolution, in which each region is anchored by an observed exemplar activation. Data is streamed from a representative corpus, and each activation either joins the region of the nearest existing exemplar, if it is within a distance threshold, or becomes a new exemplar itself. EP has a single resolution parameter: the distance threshold. The result can be understood as a Voronoi partition of activation space.

EP does not decompose activations into constituent directions or features -- in this work we show that in many of the cases where unsupervised feature discovery is useful, such as detecting concepts and understanding how model activations change in response to fine-tuning, decomposition is not necessary. Instead, regions are labelled by their \emph{dominant} features, and interpreted by a number of methods (e.g. commonalities or most important tokens in their generating prompts, logit-lens, common output tokens, etc), some of which are shown in Figure \ref{fig:headline}.

EP dictionaries are fast and cheap to build, and because every region is attached to a real activation (and a real prompt that produced it), we can follow regions across layers, checkpoints and architectures by matching exemplars. Because the dictionary grows until the stream stops producing new regions, the number of regions reflects the activation distribution of the data stream at the chosen threshold, rather than being preselected.

We begin by describing the method in detail (Section~\ref{sec:method}), then show that EP regions detect interpretable concepts on AxBench (Section~\ref{sec:meaningful}). We use EP to reveal how instruction tuning reorganises harmful prompt activations in Gemma and Llama, and show that interventions on these regions can overcome refusal (Section~\ref{sec:training-change}). We then identify regions encoding concealed information in models trained to hide a secret word (Section~\ref{sec:taboo}). Finally, we use linear probes to measure how much predictive information EP retains (Section~\ref{sec:information-loss}), and discuss limitations (Section~\ref{sec:limitations}).

The appendices cover related literature (Appendix~\ref{app:related-work}), dictionary saturation and stability (Appendix~\ref{app:properties}), correspondence with SAE features and SAEBench evaluation (Appendices~\ref{app:correspondence-table} and~\ref{app:saebench}), and extended cross-model and Taboo results (Appendices~\ref{app:crossfamily} and~\ref{app:taboo}). They also include a toy experiment illustrating the limits of finer partitioning (Appendix~\ref{app:resolution-toy}), future directions (Appendix~\ref{app:future-work}), and construction costs and dictionary build details (Appendices~\ref{app:efficiency} and~\ref{app:dictionaries}).

\section{Exemplar Partitioning}
\label{sec:method}

Clustering activations into labelled regions is a simple approach, but requires specific methodological choices in order to be tractable at realistic scale. For many applications we might want EP dictionaries to cover as much of a model's within-distribution activation space as possible, which requires a very large dataset representative of the training corpus -- millions or billions of activations. To make this possible we use online clustering, in a single streaming pass over the activations.

Online clustering requires some coordinate for each cluster to measure distance from. The usual choice is a running mean of each cluster's members (sometimes called a centroid), but we avoid this for two reasons: the mean of any set of activations need not be anything the model ever produces in reality, which makes it difficult to interpret (it may end up in an out-of-distribution region of activation space); and a running mean moves as its cluster grows, changing which later activations it attracts. We found that this often results in massive degenerate regions which swallow most of the stream as they converge on the global mean. 

For these reasons, while we do store the region means for later use, we use fixed exemplars for construction. An exemplar is a real activation from the stream, and its position is fixed when the region is created. Adding later exemplars can change the nearest-exemplar boundaries, but does not move existing representatives.

Fixed exemplars also let us compare dictionaries directly, because each exemplar is associated with the prompt that produced its activation. Any two dictionaries built from the same stream, whatever the layer, checkpoint or model, can be compared exemplar by exemplar: a prompt that created an exemplar activation in one either also created an exemplar activation in the other, or its activation was absorbed by an earlier region there (and we identify which region by simply finding the nearest one to the activation).We use this throughout the paper to map EP regions across layers, checkpoints and models.

\textbf{Calibration and Centring.} Leader clustering requires a distance threshold to determine whether a new activation joins an existing region or creates a new one. To set this, EP first samples a fixed budget of \emph{calibration} activations from the build corpus and computes within-batch pairwise distances after the centring and normalisation described below.

The distance threshold $\theta_{p}$ is set to the mean of the within-batch $p$-th percentiles of these distances, where $p$ is the resolution parameter the user chooses. Smaller $p$ produces tighter regions and a larger dictionary; larger $p$ produces coarser regions and a smaller one. In practice we sweep $p$ to build a family of dictionaries at different resolutions and compare results across the family. Reporting the threshold as a percentile rather than an absolute distance makes it comparable across layers and models: typical pairwise cosines differ, but $p_{10}$ always uses the tenth percentile within each calibration batch. This does not imply equal dictionary sizes or equivalent semantic resolution.

The same calibration activations determine a fixed centre $\mu$: we average their unit directions, normalise the result, and scale it by the mean projection of the raw activations onto that direction. Every activation is then mapped to
\[
\phi(a) = \frac{a-\mu}{\lVert a-\mu\rVert_2},
\]
before cosine distance is computed. Cosine is a natural first choice of distance metric because the working assumption in the literature is that features are directions in activation space, so the angle between two activations, not their magnitude, should determine their distance. We centre because residual-stream activations share a large common offset from the origin \citep{ethayarajh2019contextual,timkey2021rogue}, so without it the cosine similarity between any two activations is high because they are both dominated by that large shared component. Other metrics would produce different dictionaries (note that the EP method does not make a particular commitment to the distance metric or whether centring is applied); we leave that comparison to future work.

\begin{algorithm}[t]
\caption{Exemplar Partitioning batched dictionary construction}
\label{alg:ep}
\begin{algorithmic}[1]
\STATE \textbf{Input:} activation stream \(A\), centre \(\mu\), threshold
\(\theta_p\), saturation window \(W\) (default $W = 1$)
\STATE Initialize dictionary \(\mathcal{D} \leftarrow \emptyset\)
\STATE Initialize empty-batch counter \(w \leftarrow 0\)
\FOR{activation batch \(B \subset A\)}
    \STATE Convert each \(a \in B\) to \(\phi(a) = (a-\mu)/\|a-\mu\|_2\)
    \STATE Compare batch to current exemplar matrix
    \STATE Assign activations within \(\theta_p\) of their nearest exemplar
    \STATE Collect remaining activations as unmatched
    \STATE Run leader clustering within unmatched activations
    \STATE Spawn one new exemplar per unmatched leader
    \STATE Update cluster statistics for all with new members
    \STATE \(w \leftarrow w + 1\) if no new exemplars this batch, else \(w \leftarrow 0\)
    \IF{$w \geq W$}
        \STATE \textbf{break}
    \ENDIF
\ENDFOR
\STATE \textbf{return} \(\mathcal{D}\)
\end{algorithmic}
\end{algorithm}

We stream activations from the model's training data, a representative corpus, or a domain of interest, using the centre and threshold calibrated on that dataset. Each activation is centred, normalised, and assigned to its nearest exemplar if within \(\theta_p\); otherwise it becomes a new exemplar. In practice, the process is batched for efficiency (Algorithm~\ref{alg:ep}). Construction stops when a fixed saturation window passes where no new exemplars are created. We use a saturation window of one extraction batch; the released Pile dictionaries use \(\texttt{bs}=128\) and \(\texttt{ctx}=128\), so each full batch contains \(16{,}384\) activation vectors. We did not tune this. Appendix~\ref{app:efficiency} reports the resulting build budgets, and Appendix~\ref{app:dictionaries} lists the released dictionaries. The saturated dictionary size is itself a statistic of the activation distribution: under matched calibration it varies by domain and layer, and may measure how many distinct regions the model uses for a domain at a given layer and threshold rather than the domain's complexity (Appendix~\ref{app:saturation}).

Streaming order is a free parameter. In \S\ref{sec:stability} we show which regions reproduce when the stream is shuffled, and give one per-region statistic that predicts this.

During construction, we attach a number of values to each region: the \emph{exemplar} $e_i = \phi(a_{\mathrm{first},i})$, which is immutable and used for nearest-exemplar assignment; the \emph{region mean} $m_i = \operatorname{normalise}\!\bigl(\sum_{a\in C_i}\phi(a)\bigr)$, the mean of the region's members and a smoother summary of the region interior (note that the mean is not a real activation, so while it may be a better descriptor, it may also lie outside the model's activation distribution; we compare the two in later experiments); and the \emph{member coherence}
\[
c_i = \frac{\bigl\|\sum_{a\in C_i}\phi(a)\bigr\|_2}{|C_i|},
\]
which is $1$ when all members point in the same direction and decreases as the members spread out. We also store the prompt that resulted in the exemplar activation, the member count, the mean and variance of member distance to the exemplar, and optionally other statistics or member examples.

Coherence and member count together predict cross-seed reproducibility (\S\ref{sec:stability}; details in Appendix~\ref{app:stability}). The corpus centre $\mu$ is cached for inference.

\subsection{Inference}

The result is an EP dictionary: a Voronoi partition of the centred unit sphere under cosine distance, where every activation belongs to exactly one region (Figure~\ref{fig:headline}). At inference, an incoming activation is centred and normalised and assigned to its nearest exemplar. We use three \emph{readouts}: the native readout returns only the closest region, a top-$n$ readout returns the $n$ nearest, and a dense readout returns distances to every exemplar.

Note that an EP region is a region of activation space anchored by an observed activation. If we return several nearby regions, we learn that an activation is near both $A$ and $B$, but this is not an additive decomposition into $A+B$.

\subsection{Dictionary Stability}
\label{sec:stability}

Leader clustering depends on streaming order: whichever activation arrives first becomes the exemplar. We cannot remove this dependence, but we can predict, from the saved dictionary alone, which regions it affects. If $|C_i|$ is the region's member count and $c_i$ its member coherence, the statistic
\[
D_i = \log_{10}(|C_i| c_i^2)
\]
expresses a coherence-weighted effective sample size on a logarithmic scale. Across Gemma-2-2B~\citep{gemma2024} L12 dictionaries built on the Pile~\citep{gao2020pile} with five streaming seeds at $p \in \{2,4,8\}$, $D_i$ predicts region stability across seeds, measured by cosine similarity between region means paired using the Hungarian algorithm at Spearman $\rho \approx 0.68$ across a $16\times$ range in dictionary size. The highest-scoring fifth of regions averages matched cosine $0.80$--$0.83$, while the lowest-scoring fifth averages $0.59$--$0.62$ (Appendix~\ref{app:stability}).

The first-arrival exemplar and the exact Voronoi boundary will still move when the stream is shuffled. But a researcher can rank regions by $D_i$ and set the unstable ones aside before interpreting, steering, or comparing them. SAEs have the same problem, since seed alone changes the learned feature set \citep{paulo2025different}, but here the instability can be measured region by region from a single build. Appendix~\ref{app:stability} contains the full stability results.

\section{Are EP Regions Meaningful?}
\label{sec:meaningful}

On AxBench latent concept detection \citep{wu2025axbench} at Gemma-2-2B-it L20, an EP dictionary with a threshold of $p_{1}$ reaches mean AUROC of $\mathbf{0.937}$, an increase of $+0.026$ over SAE-A, the strongest dictionary-based baseline, while using $\sim10^{3}\times$ fewer construction tokens than the Gemma Scope SAE. It is also within $0.009$ of DiffMean and Probe, which fit a direction per concept from labelled activations, even though EP sees the labels only when selecting a concept region.

\subsection{AxBench Concept Detection}
\label{sec:axbench}

AxBench \citep{wu2025axbench} is a benchmark for concept detection: given a named concept, how well can an interpretability method tell apart texts that contain the concept from texts that do not? It provides $500$ concepts taken from the automatic descriptions of Gemma Scope SAE features, for example ``references to debugging processes and tools'', ``biographical information about a person'', ``terms related to root in mathematical contexts''. For each concept we have $72$ short texts that contain the concept (the positive texts); and $72$ that do not (the negative texts). Together these $144$ texts are the concept's `training set'. A separate test set contains $36$ positive texts and $36$ negative texts. A detector for a concept is any procedure that assigns a number to a text, with a higher number meaning the concept is more likely present. AxBench evaluates performance using the average area under the receiver operating characteristic curve (AUROC) over all concepts: $1.0$ means perfect separation of positive and negative texts, and $0.5$ means chance performance.

Following AxBench's SAE-A selection rule, we select one EP region to act as the detector for each concept, using that concept's training set. For a given region, we score each text by the maximum, over the text's tokens, of the cosine similarity between the token's activation and the region's representative (either its exemplar or its mean). For each concept we compute this score for every region, measure each region's AUROC on the $144$ training texts, and keep the region with the highest value. We then score the held-out test texts with that region alone, and report the test AUROC averaged over the $500$ concepts.

This selection rule matches SAE-A (\texttt{GemmaScopeSAEMaxAUC}~\citep{wu2025axbench}), but the candidate sets differ in size: SAE-A uses $\sim16$k GemmaScope features; the best performing EP dictionary, at $p_{1}$, has $5{,}796$ regions. The training data also differ: Gemma Scope SAEs are trained on activations from text drawn from the Gemma pretraining distribution \citep{lieberum2024gemmascope}, whereas the EP dictionaries here are by necessity built from public Pile activations. Other, supervised methods in AxBench fit a per-concept direction or classifier from labelled activations, while the canonical SAE baseline uses the GemmaScope feature already associated with each concept in the benchmark, without searching for another feature.

The best configuration tested, $p_{1}$ using region means, reaches mean AUROC $\mathbf{0.937}$: $+0.182$ over the canonical GemmaScope SAE row ($0.755$) and $+0.026$ over SAE-A ($0.911$), the strongest dictionary-based baseline. Every EP dictionary tested ($p_{0.5}$, $p_{1}$, $p_{2}$, $p_{4}$ and $p_{8}$) beats the canonical SAE result with both exemplar and region mean representatives.

Of the supervised methods, DiffMean and Probe fit one direction per concept from the labelled training activations and reach $0.946$; ReFT-r1, which trains an intervention per concept, reaches $0.965$. EP at $p_{1}$ (\Cref{tab:axbench}) uses the labels only to select representatives from its previously built dictionary, and scores $0.009$ below the fitted directions, while exceeding BoW ($0.931$), a supervised bag-of-words classifier, and the prompted detector ($0.921$). Thus, on this benchmark, selecting an existing region representative performs nearly as well as fitting a direction to labelled activations.

\begin{table}[t]
\caption{AxBench-aligned latent concept detection on Gemma-2-2B-it L20.
Top: EP from this paper across percentile resolutions, full
$500$ AxBench concepts. Bottom: cited leaderboard scores
\citep{wu2025axbench} on the same setting. Mean AUROC across concepts on AxBench test set, computed with AxBench's \texttt{AUCROCEvaluator} procedure. Region selection follows the SAE-A protocol.
Region count $K$ varies with the percentile threshold; finer percentiles
produce wider dictionaries with more candidate detectors per concept.
Supervision labels: \emph{labels at construction} fits the per-concept
direction or classifier on labelled training data; \emph{labels at selection} picks one element of a label-free dictionary using concept labels; \emph{paired} uses one pre-paired SAE feature per concept (feature labels embedded in dataset construction without per-concept search).
\textbf{SAE-A is the baseline with the same protocol as EP}: both select from a label-free dictionary using concept labels.}
\label{tab:axbench}
\centering
\small
\begin{tabular}{lrrr}
\toprule
 & & \multicolumn{2}{c}{Mean AUROC} \\
\cmidrule(lr){3-4}
Method (Gemma-2-2B-it L20) & $K$ & Region mean & Exemplar \\
\midrule
EP ($p_{0.5}$) & 18{,}674 & 0.930 & 0.914 \\
EP ($p_{1}$)   & 5{,}796 & 0.937 & 0.832 \\
EP ($p_{2}$)   & 2{,}037 & 0.915 & 0.842 \\
EP ($p_{4}$)   &    686  & 0.919 & 0.864 \\
EP ($p_{8}$)   &    226  & 0.889 & 0.866 \\
\midrule
Supervision (leaderboard baseline) & $K$ & \multicolumn{2}{c}{Mean AUROC} \\
\midrule
ReFT-r1 (labels at construction)        & -- & \multicolumn{2}{c}{0.965} \\
DiffMean (labels at construction)       & -- & \multicolumn{2}{c}{0.946} \\
Probe (labels at construction)          & -- & \multicolumn{2}{c}{0.946} \\
\textbf{EP $p_{1}$, region mean (labels at selection)} & 5{,}796 & \multicolumn{2}{c}{\textbf{0.937}} \\
BoW (labels at construction)            & -- & \multicolumn{2}{c}{0.931} \\
Prompt (concept text as prompt)         & -- & \multicolumn{2}{c}{0.921} \\
\textbf{SAE-A (labels at selection)}    & 16{,}384 & \multicolumn{2}{c}{0.911} \\
LAT (labels at construction, pos.\ only)& -- & \multicolumn{2}{c}{0.809} \\
SAE (paired feature, no selection)      & 16{,}384 & \multicolumn{2}{c}{0.755} \\
PCA (labels at construction, pos.\ only)& -- & \multicolumn{2}{c}{0.712} \\
\bottomrule
\end{tabular}
\end{table}

\section{Comparing Harmful Refusal in Gemma and Llama Before and After Instruction Tuning}
\label{sec:training-change}

EP allows us to compare activations across models (even those of different architectures) without any learned alignment, by matching exemplar prompts. In this experiment, we consider Gemma-2-2B-it and Llama-3.1-8B-Instruct~\citep{grattafiori2024llama3}, and use EP to see how the activations of harmful and benign prompts are structured pre- and post-instruction tuning, and how this compares across the two architectures. We can also use EP regions to intervene on refusal. At L14, moving activations away from the harmful regions and toward the benign ones leads to substantive answers on $28/32$ harmful prompts in Gemma and $29/32$ in Llama. 

To map EP regions between models of different architectures we do the following: for a region in model A, we take the prompt that resulted in its exemplar activation, and locate the region containing that prompt's activation in model B. We then take \emph{that} region's exemplar prompt, and map it back to A. A correspondence between regions is counted when this round trip returns to the starting region in A. We compare the matches against a permutation null; Appendix~\ref{app:crossfamily} contains the null procedure, full layer-by-layer results, and robustness checks.

EP dictionaries are built on final-token activations from 1,176 deduplicated harmful prompts drawn from AdvBench~\citep{zou2023advbench}, JailbreakBench~\citep{chao2024jailbreakbench}, MaliciousInstruct~\citep{huang2023maliciousinstruct}, StrongREJECT~\citep{souly2024strongreject}, and HarmBench~\citep{mazeika2024harmbench}, paired with 1,176 benign Alpaca prompts~\citep{taori2023alpaca}. All prompts use each model's chat format; dictionaries are built at $p=12$ at every layer of Gemma-2-2B, Gemma-2-2B-it, Llama-3.1-8B, and Llama-3.1-8B-Instruct. We call a region `harmful' when all its members come from harmful prompts, and `benign' when all come from benign prompts.

We find that both base models already separate some harmful prompts from benign ones by topic, though neither does so cleanly. In Gemma, the share of harmful prompts in a wholly harmful region varies between $2\%$ and $23\%$ across layers, with no relation to depth; in Llama it varies between $23$\% and $53$\% at layers 10--14 before falling back to $18\%$ at the final layer. Shuffling which prompts count as harmful while leaving the regions unchanged drops both figures to almost nothing, so this grouping is not a consequence of region size.

\begin{figure}[t]
  \centering
  \includegraphics[width=\linewidth]{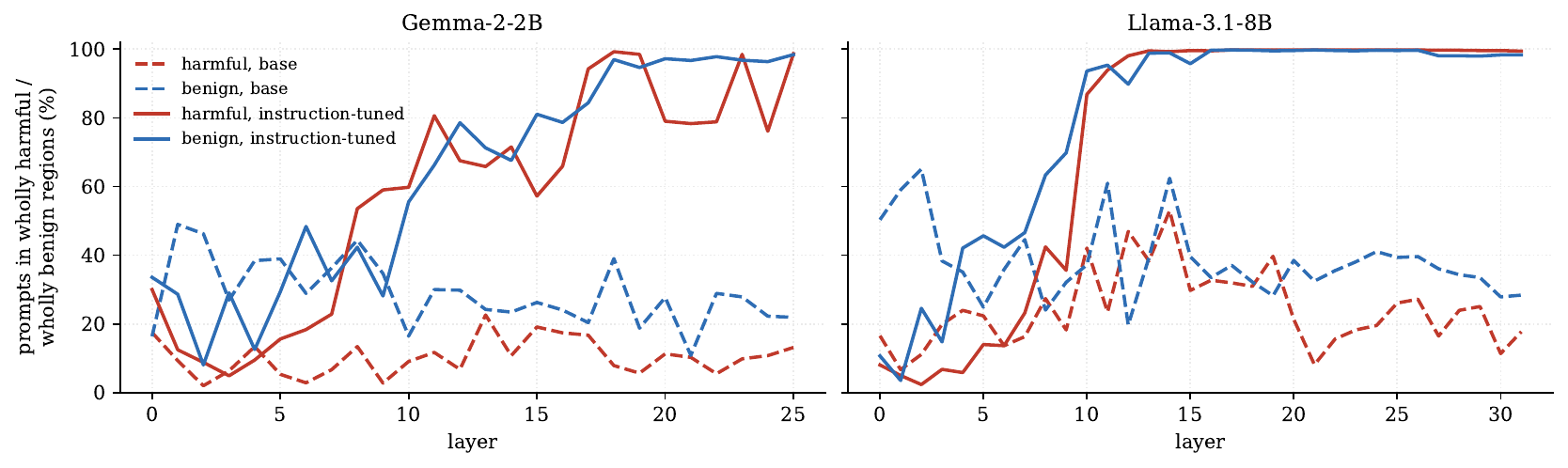}
  \captionsetup{font=small}
  \caption{Separation of harmful and benign prompts by layer. Each line is the share of harmful prompts whose final-token activation lies in a wholly harmful region (red), or of benign prompts in a wholly benign region (blue), for the base (dashed) and instruction-tuned (solid) checkpoints of Gemma-2-2B and Llama-3.1-8B. Both tuned models separate the two groups almost completely, Llama from L12 onwards and Gemma at L18; neither base model does this at any layer.}
  \label{fig:separation-by-layer}
\end{figure}

After instruction tuning, almost every harmful prompt falls in a wholly harmful region: $99\%$ in Gemma and $100\%$ in Llama at their best-separated layers. Measured at the same resolution, tuning moves Gemma's harmful prompts into a few relatively large regions and divides benign material finely, while Llama does the reverse, drawing finer distinctions within harmful content. The dictionaries have similar total sizes ($K=162$ for Gemma, $161$ for Llama), but Gemma has $37$ wholly harmful regions averaging $31.5$ prompts each and $119$ wholly benign regions averaging $9.6$; Llama has $91$ wholly harmful regions averaging $12.9$ prompts and $68$ wholly benign regions averaging $17.2$. Gemma therefore groups harmful prompts into fewer, larger regions, while Llama divides them more finely. Near-complete harmful coverage emerges at different depths: Llama places $98\%+$ of harmful prompts in wholly harmful regions from L12 onwards, while Gemma reaches $99.2\%$ at L18 before falling again (\Cref{fig:separation-by-layer}).

\begin{figure}[t]
  \centering
  \includegraphics[width=\linewidth]{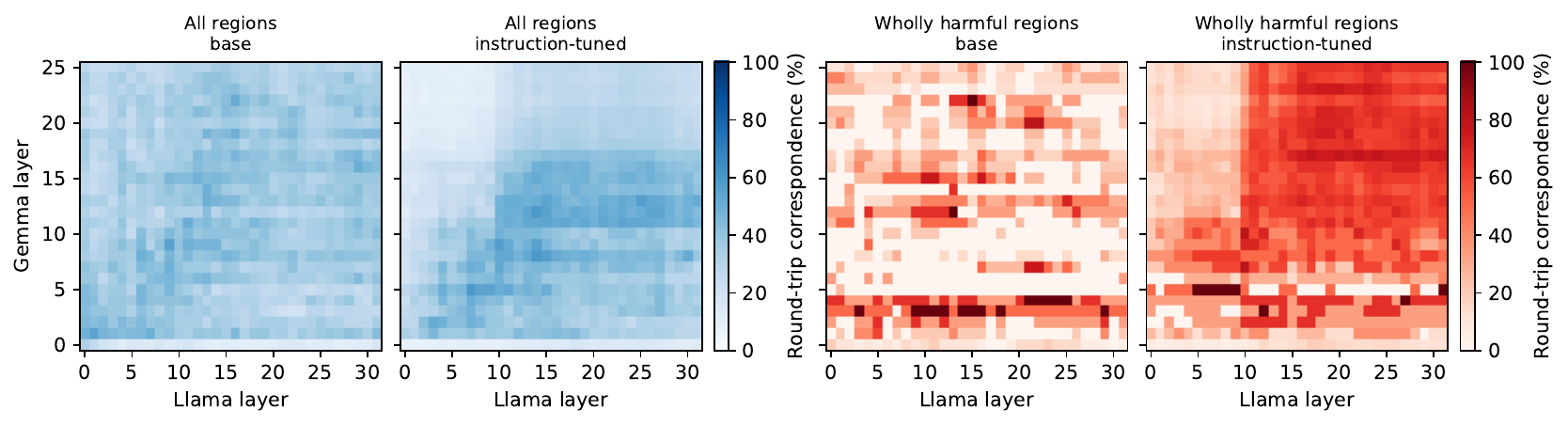}
  \captionsetup{font=small}
  \caption{Round-trip correspondence between Gemma and Llama regions at every layer pair, for all Gemma regions (blue) and wholly harmful Gemma regions (red), before and after instruction tuning. Each cell is the share of Gemma regions whose exemplar prompt maps to a Llama region whose exemplar prompt returns to the starting region. Across all regions the mean is $34.5\%$ before tuning and $32.4\%$ after; across wholly harmful regions it rises from $18.7\%$ to $43.9\%$.}
  \label{fig:crossfamily-grid}
\end{figure}

Instruction tuning increases the cross-model correspondence between harmful regions (\Cref{fig:crossfamily-grid}): averaged over all layer pairs, the share of Gemma harmful regions that round-trip through Llama rises from $19\%$ to $44\%$, while across all regions, correspondence actually falls slightly from $34.5\%$ to $32.4\%$. At the layers where both tuned models best separate harmful and benign prompts, $68\%$ of Gemma's harmful regions round-trip through Llama. Two simpler explanations for the separation, dictionary size and dataset-specific wording, are ruled out in Appendix~\ref{app:crossfamily}: the base--tuned gap persists at matched $K$, and a TF--IDF $k$-means baseline with the same number of clusters places only $27\%$ of harmful prompts in wholly harmful clusters.

Membership in a wholly harmful region does not guarantee refusal. Of the harmful prompts that the models comply with, $75/78$ for Llama and $32/34$ for Gemma still fall in a wholly harmful region. All 68 prompts that Gemma refuses and Llama answers fall in a wholly harmful Llama region. Thus, harmful/benign separation persists even when refusal fails. This suggests that these failures occur after harmfulness is represented, rather than because the model fails to distinguish harmful from benign inputs. Across the tuned checkpoints, $23/37$ Gemma harmful regions and $52/91$ Llama harmful regions are fully refused. This is consistent with supervised evidence that harmfulness and refusal are encoded separately \citep{zhao2025harmfulness}.

\subsection{Causal Intervention}
We can intervene on refusal using region means. At a chosen layer, we remove each activation's component in the harmful-region mean span and replace it with the average benign region mean's component in that span. This \emph{swap} has no tuned strength parameter. This extends affine concept editing from one direction to a span \citep{marshall2024affine}; refusal can occupy several directions \citep{wollschlager2025geometry}. We exclude the evaluation prompts from the region means used for this intervention.

We test Gemma layers 8, 14, 18, and 24, and Llama layers 10, 14, 17, and 31. At each layer, we generate 256 token completions for 32 harmful prompts and ask Claude Opus 4.8 whether each response provides the requested harmful content. Intervening at L14, compliance rises from $0.00$ to $0.88$ in Gemma-2-2B-it and from $0.00$ to $0.91$ in Llama-3.1-8B-Instruct. Compliance rises to only $0.12$--$0.22$ at the other tested layers, except Llama's final layer, where almost all generations are empty.

Removing the harmful-region component at L14 without adding the benign component results in compliance rates of $0.00$ in Gemma and $0.34$ in Llama, compared with $0.88$ and $0.91$ for the full swap. Projecting off a benign-region span of the same dimension, or an equal number of random directions, results in $0.00$ compliance in both models (\Cref{tab:region-intervention-controls}). Replacing the removed component with the benign component is therefore necessary for the full effect. The swap is also less effective at the layers with the clearest harmful/benign separation: compliance reaches only $0.22$ at Gemma L18 and $0.16$ at Llama L17, so the degree of separation does not predict the effect of the intervention. Full method and results are in Appendix~\ref{app:region-intervention}.

\section{Finding a Secret Word in Model Activations}
\label{sec:taboo}

The Taboo testbed of \citet{cywinski2025secret} contains 21 versions of Gemma-2-9B-it, each with a different target word, such as \emph{gold} or \emph{book}. Each version is fine-tuned to give clues about that word and answer questions about it, while avoiding the word itself in its replies at all costs. We test whether EP can identify that hidden word in the model's activations.

For each model, we build a $p=12$ dictionary from 500 hint-game transcripts at layer 32. We use only the two fixed chat-format tokens at the start of each assistant turn, so the sampled tokens themselves never contain the secret word. We then run the base model on the same transcripts and collect its activations at the same positions. To find regions that might encode the hidden word (i.e. those in this data that newly appear after fine-tuning), we list the regions that contain fine-tuned activations but no base-model activations at the sampled positions. These candidate regions are selected without using the hidden word.

To label each candidate region, we rank all vocabulary tokens by the cosine similarity between their embedding vectors and the region mean. The highest-ranked tokens provide a way to interpret the region. In 17 of the 21 models, at least one candidate region ranks the hidden word among its top four tokens; in 12, it ranks first. The regions for \emph{book} and \emph{song} rank related words highly, such as \emph{novels} and \emph{musicians}. Including these two cases, 19 of 21 models have a candidate region whose highest-ranked tokens are recognisably related to the hidden word. We find no such region for \emph{rock} or \emph{salt}, including at the other layers we tested. Appendix~\ref{app:taboo} contains the full results.

\section{Information Loss}
\label{sec:information-loss}

We find that a linear classifier trained to predict text labels does so almost as well from how often a text's tokens fall into EP regions, as it does from the activations themselves. At L12 of Gemma-2-2B, a classifier restricted to one region of a $p_{10}$ dictionary reaches $98\%$ of the accuracy of one restricted to a single activation dimension, and a classifier on the regions of a $p_{1}$ dictionary reaches $90.5\%$ of the accuracy of a full activation probe.

We measure this with SAEBench's sparse probing suite \citep{karvonen2025saebench}, which trains linear classifiers to predict text attributes such as profession, sentiment, news topic, and natural or programming language, and averages accuracy over the tasks. Each classifier reads one vector per text: using EP, this vector contains the fraction of the text's tokens assigned to each region; vs the mean of the raw activations over the text's tokens. SAEBench fits one classifier on the entire vector, and one on the single element whose mean differs most between the positive and negative training texts. EP performance increases with dictionary size: a classifier on the entire vector reaches $82\%$ of the full activation probe at $p_{10}$ ($K=203$), $85\%$ at $p_{4}$ ($K=1{,}193$), $89\%$ at $p_{2}$ ($K=5{,}129$), and $90.5\%$ at $p_{1}$ ($K=20{,}295$).

\section{Limitations} 
\label{sec:limitations}

Intuitively, if we imagine any single activation to be made up of a weighted sum of features, two activations are close together in activation space when they share their heavily weighted features. So at coarse resolution an EP region collects the activations that share their heaviest feature, and as resolution increases the regions split along progressively lighter features. Suppose the activation at a colour word carries a colour feature with a large weight, a smaller weight on a feature for the language of the sentence, and a smaller weight still on its grammatical role. At coarse resolution, EP might return one region per colour, with French and English colour words together. At finer resolution each colour region separates by language, and at finer still by grammatical role, with the dictionary growing multiplicatively each time. EP never recovers the colour, language and grammar features alone; it records which activations agree on the features that carry the most weight. A feature whose weight is always small relative to the other features in the same activation only gets regions of its own once the resolution is fine enough to separate everything above it first. Appendix~\ref{app:resolution-toy} illustrates this behaviour in a colour-pair toy: finer resolution separates colour combinations without isolating the background colour.

As with SAEs, we label EP regions using a sample of their members, via a number of methods: the logit lens; highly attributed input tokens; common firing or output tokens; an LLM asked what the member prompts have in common. Each of these labels a region by the dominant feature that the labelling method can describe at that resolution.

Features that dominate some sampled activations may be easier to identify than features that are always weaker than their neighbours. Dominance alone does not guarantee recovery: sampling, geometry, resolution, and the labelling method all matter. Take a feature for grammatical role that always appears alongside a weightier feature for the word itself. Finer resolution may separate grammatical roles within each word, while labels based on the shared content of a region may still report the word. Labelling methods that look at a region's neighbours might do better: subtract the nearest neighbours from the exemplar or region mean and logit-lens the remainder, or ask an LLM what differs between neighbouring regions. We leave this for future work.

EP dictionaries depend to a large extent on streaming order; we discuss this briefly in \S\ref{sec:stability} and give a metric for identifying unstable regions in a single dictionary. Small dense regions are stable, because any exemplar drawn from them is close to every other member, but for large or uniformly populated regions of activation space the instability may be irreducible.

Appendix~\ref{app:future-work} outlines possible extensions; Appendix~\ref{app:related-work} discusses related work.

\bibliographystyle{plainnat}
\bibliography{references}

\appendix

\section{Related Work}
\label{app:related-work}

\paragraph{Unsupervised interpretability.}
Recent surveys identify two open problems in mechanistic interpretability: discovering features without labels at scale, and evaluating those features \citep{bereska2024review,sharkey2025openproblems}. EP is a discovery method: it partitions all of activation space without concept labels; we then use AxBench and Taboo to test the resulting regions against labelled concepts and deliberately hidden words.

\paragraph{Clustering, vector quantisation, and exemplar memory.}
EP is leader clustering \citep{hartigan1975clustering} applied to centred, normalised LM activations. Our implementation makes two choices: the threshold is calibrated as a percentile of pairwise distances sampled from the build corpus, and construction stops when new regions stop appearing for a specified number of batches. Vector quantisation \citep{lloyd1982least,vandenoord2017vqvae} differs in learning representatives and fixing codebook size in advance. Activation Atlas and related visualisation pipelines have aggregated neural activations for interpretability \citep{carter2019activation}, and kNN-LM uses stored activation exemplars for language modelling \citep{khandelwal2020knnlm}. EP uses stored exemplars to build unsupervised feature dictionaries, grouping nearby activations to keep the dictionary small enough to inspect.

\paragraph{Sparse autoencoders and their evaluation.}
Sparse autoencoders build on sparse coding and dictionary learning \citep{olshausen1996sparse}, and are motivated in interpretability by superposition: a model may represent more features than it has activation dimensions \citep{elhage2022toy}. SAEs have recovered interpretable features in residual streams and MLPs \citep{cunningham2023sparse,bricken2023monosemanticity}, and have scaled to millions of features \citep{templeton2024scaling}; TopK, JumpReLU, and related variants improve the reconstruction-sparsity tradeoff \citep{gao2024topk,rajamanoharan2024jumprelu}. Gemma Scope provides the pretrained SAEs used as our main comparison \citep{lieberum2024gemmascope}. Recent work studies how geometric relationships between SAE directions relate to their meanings \citep{li2025geometry}; Appendix~\ref{app:correspondence-table} compares EP representative directions with SAE features by overlap of their highest-scoring token activations. EP differs from SAEs in that it partitions observed activations instead of learning a linear basis under a reconstruction objective. We use an adapter to evaluate EP with SAEBench \citep{karvonen2025saebench}, and test concept detection with AxBench \citep{wu2025axbench}.

\paragraph{Local geometry and cross-model comparison.}
Mixtures of factor analysers also replace global directions with local regions, but learn Gaussian centroids and a covariance for each region; EP fits no parameters and anchors each region on an observed activation \citep{shafran2026regions}. Relative representations similarly compare latent spaces using the same input examples as reference points \citep{moschella2023relative}. SAE dictionaries can be compared through activation correlation, Hungarian matching, or parameter matching \citep{lan2025universality,paulo2025different,balagansky2024permutability}. Crosscoders instead learn a shared dictionary for base and fine-tuned models, and recent work extends them across architectures \citep{lindsey2024crosscoders,minder2025crosscoders,jiralerspong2026crossarchitecture}. EP's round trip through shared prompts needs neither a learned shared basis nor parameter alignment. The Taboo testbed evaluates whether white-box methods can elicit knowledge a model uses but does not state \citep{cywinski2025secret}.

\paragraph{Probing and steering.}
Linear probes test whether a specified label can be predicted from activations \citep{alain2017understanding,belinkov2022probing}; activation steering and refusal-direction work use activation differences or learned directions to control behaviour \citep{turner2023activation,panickssery2023steering,zou2023representation,arditi2024refusal}. Harmfulness and refusal have also been separated with supervised directions \citep{zhao2025harmfulness}; our dictionaries show a similar separation without using labels during construction. The linear-representation hypothesis motivates directional readouts, but multi-dimensional features and unreliable steering show where a single direction can fail \citep{engels2025multidimensional,braun2026unreliability}.

\section{Properties of the EP Dictionary}
\label{app:properties}

\subsection{Domain Saturation}
\label{app:saturation}

A trained SAE has its dictionary size set in advance through an expansion factor, the ratio of dictionary size to activation dimension. EP grows the dictionary until the activation stream stops producing new regions at the chosen resolution, so the final size and growth curve describe the sampled distribution under that threshold and stopping rule. We compare three input distributions using the same calibration at each layer: GSM8K (math), HumanEval + MBPP (code), and HH-RLHF (chat), all on Gemma-2-2B layers L4, L12, and L20 with the Pile calibration at $p = 8$ and a 10M-activation cap per domain.

\begin{figure}[t]
  \centering
\includegraphics[width=\linewidth]{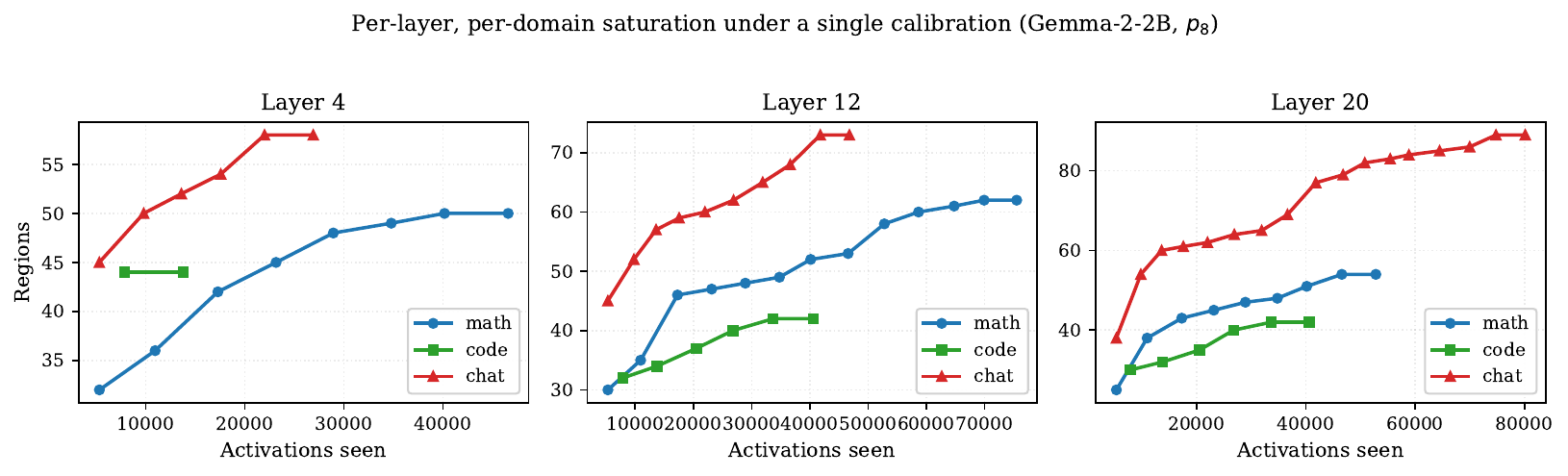}
  \caption{Dictionary growth for each domain and layer, using a Pile-calibrated
threshold at each layer ($p=8$, Gemma-2-2B). Each curve shows the running count of regions discovered as activations from one domain stream in. Chat saturates at the largest count and grows with depth; code is the flattest and saturates earliest; math peaks at L12. Final saturated counts agree with \Cref{tab:saturation}.}
  \label{fig:saturation}
\end{figure}

The three distributions saturate at different sizes, and the difference varies across layers (\Cref{fig:saturation,tab:saturation}). Chat grows monotonically with depth ($58, 73, 89$), code is essentially flat ($44, 42, 42$), and math is non-monotonic and peaks at L12 ($50, 62, 54$). The chat-to-code size ratio rises from $1.3$ at L4 to $2.1$ at L20. An earlier L12 run at $p = 10$ with a $200$k-activation cap gave the same ordering (chat $72$, math $57$, code $30$), although only code saturated there. All nine reported $p = 8$ builds reached saturation at \texttt{sat\_window=1}.

``Saturated'' is defined at the batch level: one full extraction batch with no new regions. With per-position extraction, the released Pile dictionaries use $\texttt{batch\_size} = 128$ prompts $\times$ $\texttt{ctx} = 128$ tokens, i.e.\ $16{,}384$ activation vectors producing zero new regions. This is a stopping rule on the observed stream, not a guarantee that future inputs cannot create regions. We have not measured how dictionary size changes with a longer saturation window; shorter domain sequences also produce fewer activations per extraction batch.

The number of activations processed varies by domain because each build stops at saturation. At L4, math uses $46{,}573$ activations and code uses $13{,}755$.

\begin{table}[t]
\caption{Final dictionary sizes for each domain and layer, using fixed Pile calibration
(Gemma-2-2B, $p=8$, $\texttt{sat\_window}=1$, 10M activation cap). Every build saturated.}
\label{tab:saturation}
\centering
\small
\begin{tabular}{llrr}
\toprule
Layer & Domain (source) & Activations seen & Regions \\
\midrule
L4  & math (GSM8K)              & 46{,}573  & 50 \\
L4  & code (HumanEval + MBPP)   & 13{,}755  & 44 \\
L4  & chat (HH-RLHF)            & 26{,}897  & 58 \\
\midrule
L12 & math                      & 75{,}530  & 62 \\
L12 & code                      & 40{,}586  & 42 \\
L12 & chat                      & 46{,}804  & 73 \\
\midrule
L20 & math                      & 52{,}800  & 54 \\
L20 & code                      & 40{,}586  & 42 \\
L20 & chat                      & 80{,}144  & 89 \\
\bottomrule
\end{tabular}
\end{table}

\subsection{Dictionary Stability Details}
\label{app:stability}

Changing the stream order changes which activations become exemplars and where the Voronoi boundaries fall. Regions therefore need not be identical across seeds. We want to know whether the regions correspond to structure in the activations that any sufficiently long build would find, or are artefacts of streaming order.

\paragraph{Setup and definition.}
We build $S$ dictionaries on the same data with different streaming seeds, identical calibration $(\mu, \theta)$, and the same token budget and saturation stopping rule. For each pair $(A, B)$ of dictionaries, we use the Hungarian algorithm to pair regions by cosine similarity between their means. Region means are less sensitive to stream order than exemplars. The \emph{stability} of region $i$ in dictionary $A$ is then the average matched cosine over dictionaries built with the other seeds:
\[
\mathrm{stab}(i; A) = \mathbb{E}_{B \neq A}\bigl[\cos\bigl(m^{A}_{i},\, m^{B}_{j^{*}(i)}\bigr)\bigr],
\]
where $j^{*}(i)$ is the Hungarian-matched region in $B$. Higher matched cosine means region $i$ is consistently found by independent rebuilds.

The local exemplar-mean cosine $s_i = e_i \cdot m_i$ does not predict stability (\Cref{tab:stability-correlations}). The reason is geometric: in a region of uniform density, any seed produces a representative exemplar by symmetry, but the boundaries different seeds draw within that region are arbitrary, so the regions are unstable even when their exemplars are representative.

\paragraph{A predictor.}
A region is a set of nearby activations, and its identity is their mean. Whether that mean survives a rebuild with a different streaming order depends on two properties that a single saved dictionary already records:
\begin{itemize}
\item \textbf{Member count $|C_i|$.} Averaging more members reduces the noise in the mean.
\item \textbf{Coherence $c_i$.} If the members are spread across the region, the mean is poorly determined; if they are tightly grouped, it is well determined.
\end{itemize}
Neither alone is enough. A region with many members that disagree, such as a grammatical region that absorbs many kinds of token, has a poorly determined mean despite its size. A region with five members in tight agreement has a well determined mean but too few samples. We combine these properties by multiplication: member count weighted by squared coherence is large only when a region has many members \emph{and} those members agree.

The statistic
\[
D_i \;=\; \log_{10}(|C_i| \cdot c_i^{2})
\]
combines them. Both inputs are stored on every region as scalars (\Cref{sec:method}), so $D_i$ can be computed from a single saved build. Squaring $c_i$ is the natural choice when averaging unit vectors: disagreement among members compounds in the mean, so coherence enters the precision of the mean quadratically rather than linearly. Taking the logarithm compresses the wide range of member counts, and a unit step in $D_i$ is roughly a tenfold gain in the expected precision of the mean.

\paragraph{Empirical test (Gemma-2-2B L12 Pile, 5 seeds at each of
$p \in \{2, 4, 8\}$).} We build $5$ dictionaries per resolution with streaming seeds $\{0, 1, 2, 3, 4\}$, identical calibration, and a $50\,M$ construction-token budget capped by \verb|sat_window=1|. These are separate five-seed builds from the seed-0 dictionaries used for detection and correspondence; the table reports mean $K$ across seeds. For each region in each dictionary we average its available Hungarian matched cosines across the four other-seed dictionaries to obtain $\mathrm{stab}(i; A)$. Unequal dictionary sizes can leave regions unmatched; we omit missing matches from the average and exclude regions with no matches. \Cref{tab:stability-correlations} reports Spearman correlations between each single-dictionary predictor and this per-region stability outcome.

\begin{table}[t]
\caption{Spearman correlation $\rho$ between statistics computed from one dictionary
and region stability across seeds ($\mathrm{stab}(i; A)$). Each row is one resolution; column $D_i = \log_{10}(|C_i| c_i^{2})$ is the proposed metric. The $s_i$ column measures whether an exemplar's proximity to its region mean predicts stability.}
\label{tab:stability-correlations}
\centering
\small
\begin{tabular}{rrrrrrrr}
\toprule
$p$ & mean $K$ & regions & $s_i$ & $c_i$ & $\log_{10}|C_i|$ &
$\log_{10}(|C_i| c_i)$ & $D_i = \log_{10}(|C_i| c_i^{2})$ \\
\midrule
$2$ & $4933.6$ & $24{,}614$ & $-0.27$ & $+0.30$ & $+0.65$ & $+0.66$ & $\mathbf{+0.68}$ \\
$4$ & $1236.6$ & $\phantom{0}6{,}177$  & $-0.24$ & $+0.36$ & $+0.65$ & $+0.67$ & $\mathbf{+0.69}$ \\
$8$ & $302.8$  & $\phantom{0}1{,}511$  & $-0.18$ & $+0.43$ & $+0.62$ & $+0.65$ & $\mathbf{+0.67}$ \\
\bottomrule
\end{tabular}
\end{table}

$D_i$ reaches Spearman $\rho \approx +0.68$ at every resolution across a $16\times$ range in dictionary size, and the squared form ranks regions slightly better than the unsquared $\log_{10}(|C_i| c_i)$ at every resolution. \Cref{tab:stability-quintiles} divides regions into five equally sized groups ranked by $D_i$ and reports mean stability for each group.

\begin{table}[t]
\caption{Mean cross-seed stability for five equally sized groups ranked by $D_i$ at each resolution. Q1 has the lowest $D_i$ and Q5 the highest. Top-quintile regions average matched cosine $\sim 0.81$; bottom-quintile regions average $\sim 0.60$. Gaps are differences of the displayed rounded means.}
\label{tab:stability-quintiles}
\centering
\small
\begin{tabular}{rrcccccr}
\toprule
$p$ & regions & Q1 & Q2 & Q3 & Q4 & Q5 & gap (Q5 $-$ Q1) \\
\midrule
$2$ & $24{,}614$ & $0.62$ & $0.70$ & $0.74$ & $0.78$ & $\mathbf{0.83}$ & $+0.21$ \\
$4$ & $\phantom{0}6{,}177$  & $0.59$ & $0.68$ & $0.72$ & $0.76$ & $\mathbf{0.82}$ & $+0.23$ \\
$8$ & $\phantom{0}1{,}511$  & $0.59$ & $0.66$ & $0.71$ & $0.74$ & $\mathbf{0.80}$ & $+0.21$ \\
\bottomrule
\end{tabular}
\end{table}

\paragraph{Calibration vs ranking.}
$D_i$ ranks regions reliably across seeds, but it is not a calibrated estimate of matched cosine. The argument above accounts for sampling noise in the region mean. Movement of the region boundaries also affects the absolute cosine similarity. When the exemplar moves, the region's boundary moves with it, so matched regions are samples from two slightly different regions of activation space, not two samples of the same one. In our data this boundary shift is roughly uniform across regions, so it lowers every matched cosine without changing the ranking. Converting $D_i$ into a calibrated probability of reproduction would require a model of that shift. For ranking, only the Spearman correlation matters, and it is about $0.68$ at every resolution.

A researcher with one saved dictionary can therefore compute $D_i$ from $(|C_i|, c_i)$ alone, rank the regions, and keep only the top ones before further analysis, without building a second dictionary.

\section{Direction Correspondence to Gemma Scope}
\label{app:correspondence-table}

\begin{figure}[htbp]
  \centering
\includegraphics[width=\linewidth]{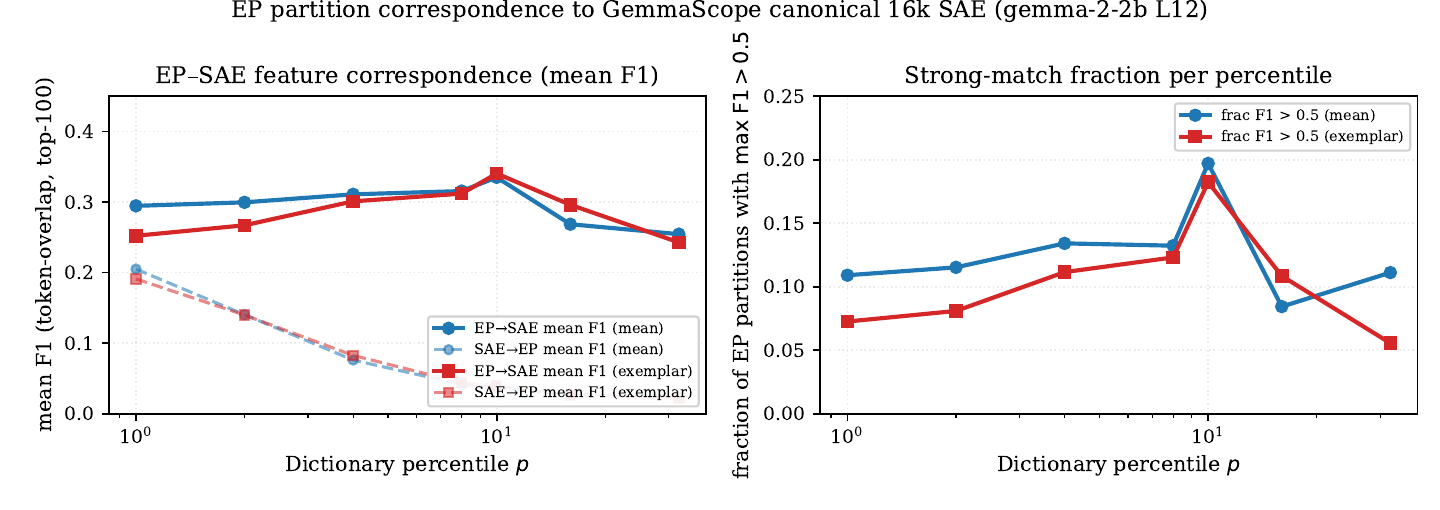}
  \caption{EP-direction correspondence to GemmaScope canonical 16k SAE on
Gemma-2-2B L12, swept across percentile resolutions. Left: mean F1 per direction. Right: fraction of EP regions with a strong (F1 $> 0.5$) SAE match. The EP$\rightarrow$SAE mean F1 peaks around $p_{10}$ ($\sim$0.34). The fraction with a strong match also peaks around $p_{10}$, where roughly one in five EP regions has a strong SAE counterpart.}
  \label{fig:compare-sae}
\end{figure}

\Cref{tab:correspondence} and \Cref{fig:compare-sae} compare EP representative directions with features of the GemmaScope canonical 16k SAE on Gemma-2-2B L12. For each EP representative, we rank token activations by centred cosine similarity; for each SAE feature, we rank them by the SAE's feature activation. We retain up to 100 positive-scoring token activations on each side. Their overlap score, $F_1$, is twice the number of shared token activations divided by the sum of the two set sizes. EP scores are computed for every representative, irrespective of which region a token belongs to. This therefore measures correspondence between direction scores, not overlap between region membership and SAE firing sets. Low correspondence under this test does not establish that either method misses concepts found by the other.

\begin{table}[htbp]
\caption{Direction correspondence to GemmaScope canonical 16k SAE on
Gemma-2-2B L12 (top-$k = 100$ activating tokens per feature, $300$--$500$k tokens scored, at least $20$ activating tokens required per SAE feature). EP$\rightarrow$SAE matches each EP region to its best SAE feature; the strong-match fraction is the share of regions whose best match has $F_1 > 0.5$ (max F1 across the sweep is $0.98$). ``SAE caught'' counts SAE features whose best EP match has $F_1 > 0.5$, as a fraction of those eligible (active in $\geq 20$ scoring tokens); the absolute count rises sharply with $K$ because a larger dictionary offers more possible matches for SAE features. Each percentile has two rows, using exemplars and region means from the same dictionary.}
\label{tab:correspondence}
\centering
\small
\setlength{\tabcolsep}{4pt}
\begin{tabular}{lrlrrr}
\toprule
Dict & $K$ & Representative & EP$\rightarrow$SAE mean F1 & strong frac & SAE caught ($F_1 > 0.5$) \\
\midrule
$p_{1}$  & 20{,}295 & mean     & 0.295 & 0.109 & $1{,}091\,/\,16{,}338$ (6.7\%) \\
$p_{1}$  & 20{,}295 & exemplar & 0.252 & 0.073 & $631\,/\,16{,}342$ (3.9\%) \\
$p_{2}$  &  5{,}129 & mean     & 0.300 & 0.115 & $531\,/\,16{,}314$ (3.3\%) \\
$p_{2}$  &  5{,}129 & exemplar & 0.267 & 0.081 & $352\,/\,16{,}329$ (2.2\%) \\
$p_{4}$  &  1{,}193 & mean     & 0.311 & 0.134 & $175\,/\,16{,}121$ (1.1\%) \\
$p_{4}$  &  1{,}193 & exemplar & 0.301 & 0.111 & $152\,/\,16{,}217$ (0.9\%) \\
$p_{8}$  &      325 & mean     & 0.316 & 0.132 & $36\,/\,14{,}145$ (0.25\%) \\
$p_{8}$  &      325 & exemplar & 0.312 & 0.123 & $60\,/\,15{,}189$ (0.40\%) \\
$p_{10}$ &      203 & mean     & 0.335 & 0.197 & $39\,/\,13{,}959$ (0.28\%) \\
$p_{10}$ &      203 & exemplar & \textbf{0.340} & 0.182 & $44\,/\,14{,}908$ (0.30\%) \\
\bottomrule
\end{tabular}
\end{table}

\section{SAEBench Comparison}
\label{app:saebench}

We evaluate EP through SAEBench's encoder/decoder interface \citep{karvonen2025saebench}, using the Gemma-2-2B L12 $p_{10}$ dictionary with $203$ regions. We test both unit exemplar directions and unit region-mean directions as representatives $b_j$. Exemplar assignment is the original EP partition; nearest-mean assignment is an alternative readout and can assign tokens differently.

\paragraph{Adapter.}
For an activation $a$, let $x=a-\mu$ and $j^*=\arg\max_j x\cdot b_j$. Sparse probing uses a binary code $z_j=\mathbf{1}[j=j^*]$. The reconstruction and RAVEL evaluations instead retain a projection amplitude:
\[
z_j=\mathbf{1}[j=j^*]\max(x\cdot b_{j^*},0),
\qquad \hat a=\mu+\sum_j z_j b_j.
\]
This amplitude uses the unnormalised centred activation, so it preserves information about magnitude that the binary code discards. These two readouts should not be interpreted as the same compression scheme.

\paragraph{Probing setup.}
\Cref{tab:saebench} reports the SAEBench results. SAEBench trains linear classifiers to distinguish texts with a particular label from texts without it, including profession, sentiment, product category, news topic, and natural or programming language. It averages feature activations over a text's tokens before fitting the probe. For binary EP codes, each coordinate is therefore the fraction of tokens assigned to that representative. For the raw-activation reference, each coordinate is the average value of one activation dimension. The unrestricted probe uses all coordinates; top-$k$ probing selects the $k$ coordinates with the largest absolute difference between positive and negative training-set means and fits a probe on those coordinates. In particular, ``top-1'' means a one-coordinate probe, not the probability that a multiclass classifier's highest-ranked answer is correct. Scores are averaged over the benchmark's classification tasks.

\texttt{IdentitySAE} passes the raw activations through the same interface. SAEBench also reports an LLM-feature reference. The two references have the same representation, though their reported top-$k$ scores differ slightly. We use IdentitySAE for the numerical comparisons in the main text.

\begin{table}[t]
\caption{SAEBench results at Gemma-2-2B L12 $p_{10}$. Sparse probing uses binary region indicators; core and RAVEL use the adapter that retains projection amplitudes described above. ``Test acc'' uses all coordinates; top-$k$ uses $k$ selected coordinates. Core measures reconstruction and model-output preservation; RAVEL measures the ability to change a target attribute while preserving others.}
\label{tab:saebench}
\centering
\small
\begin{tabular}{lrrrr}
\toprule
Module / metric & EP mean & EP exemplar & IdentitySAE & LLM ref \\
\midrule
\texttt{core}: KL-div score              & \textbf{0.540} & 0.469 & --    & --    \\
\texttt{core}: CE-loss score             & \textbf{0.524} & 0.455 & --    & --    \\
\texttt{core}: explained variance        & \textbf{0.531} & 0.496 & --    & --    \\
\texttt{core}: cossim                    & \textbf{0.703} & 0.652 & --    & --    \\
\texttt{core}: $\ell_{0}$ / $\ell_{1}$   & 1 / 52.3 & 1 / \textbf{32.5} & --    & --    \\
\midrule
\texttt{sparse\_probing}: test acc       & 0.773 & \textbf{0.786} & 0.959 & 0.959 \\
\texttt{sparse\_probing}: top-1          & \textbf{0.653} & 0.641 & 0.652 & 0.664 \\
\texttt{sparse\_probing}: top-2          & 0.672 & \textbf{0.674} & 0.720 & 0.723 \\
\texttt{sparse\_probing}: top-5          & 0.707 & \textbf{0.708} & 0.780 & 0.780 \\
\midrule
\texttt{ravel}: disentanglement          & 0.276 & \textbf{0.290} & --    & --    \\
\texttt{ravel}: cause                    & \textbf{0.060} & 0.051 & --    & --    \\
\texttt{ravel}: isolation                & 0.492 & \textbf{0.529} & --    & --    \\
\bottomrule
\end{tabular}
\end{table}

The binary code maps each $2{,}304$-dimensional activation to one of $203$ indices, requiring at most $\log_2 203\approx7.7$ bits per token before text-level averaging. A one-coordinate probe on exemplar assignments scores $0.6409$, compared with $0.6521$ for IdentitySAE: a gap of $1.12$ percentage points. Using all coordinates results in $0.786$ versus $0.959$, a much larger gap. The region-mean assignment results in a one-coordinate score of $0.6527$.

Finer dictionaries preserve more information for the unrestricted probe. Its accuracy relative to IdentitySAE rises from $82\%$ at $p_{10}$ ($K=203$) to $85\%$ at $p_4$ ($K=1{,}193$), $89\%$ at $p_2$ ($K=5{,}129$), and $90.5\%$ at $p_1$ ($K=20{,}295$), all using binary exemplar assignments. At $p_2$, the one-coordinate score is $0.6606$. These results concern prediction of the benchmark labels; they do not imply that the code preserves the same proportion of the activation's information. The core and RAVEL scores separately show the limitations of reconstructing or editing activations through a single representative.

\section{Cross-Family Correspondence}
\label{app:crossfamily}

\Cref{fig:crossfamily-grid} reports the full layer-by-layer round-trip grid using the 100\%-harmful criterion. White cells have zero correspondence; every Gemma layer has at least one wholly harmful region in both checkpoints. High correspondence between later layers appears only after instruction tuning; the all-region grid does not show the same increase. To estimate correspondence expected by chance, we randomly re-pair prompts between Gemma and Llama while leaving both sets of regions unchanged, then recount the completed round trips. At Gemma L18 and Llama L17, 25 of 37 Gemma regions return to themselves, against a null mean of 2.89 and maximum of 11 in 2,000 draws ($p<0.0005$). Repeating the comparison across Llama layers L17--26 results in 22--25 matches, while the null maximum is at most 12.

Dictionary size does not explain the result by itself. Across all 116 layers, harmful coverage and $K$ correlate at $r=0.78$, but the difference between base and tuned models persists at similar dictionary sizes: Gemma base L13 ($K=49$, $23\%$ coverage) against Gemma-it L8 ($K=57$, $54\%$), and Llama base L13 ($K=59$, $38\%$) against Llama-it L10 ($K=65$, $87\%$). The variation in Gemma's harmful coverage at later layers is also not explained by dictionary size: $K$ stays between 193 and 261 from L20--24 while harmful coverage moves between $76\%$ and $99\%$.

Nor does dataset-specific wording explain it. Because harmful and benign prompts come from different datasets, the separation could reflect surface vocabulary rather than harmfulness. To gauge how much simple lexical clustering can explain, we run a word-and-bigram TF--IDF $k$-means baseline with 161 clusters, matching the number of EP regions in Llama-it's best-separated layer (Gemma-it has 162). It places $27.0\%$ of harmful prompts in wholly harmful clusters, well below EP's $99$--$100\%$.

\subsection{Region Intervention}
\label{app:region-intervention}

We use dictionaries calibrated separately for each model and layer at $p=12$. They are built without harmful/benign labels; afterwards, we use the labels to select regions containing only harmful or only benign prompts. For this intervention, each selected region is represented by the arithmetic mean of its raw activations, recomputed after excluding evaluation prompts. These means retain magnitude, unlike the normalised direction summaries used elsewhere. At a given layer, let $c$ be the calibration centre, $B$ an orthonormal basis for the span of the harmful region means in centred space, and $\bar b$ the average of the benign region means. The swap maps an activation $x$ to $x-BB^\top(x-c)+BB^\top(\bar b-c)$. It removes the component along the harmful span and puts back the benign regions' mean position in that span, without fitting an intervention strength.

We test four layers per model on 32 harmful prompts per layer drawn from the 2,352-prompt construction corpus. We also select 32 benign prompts per layer so that all 64 evaluated prompts can be excluded from the harmful and benign region means before we compute the swap; none produced a region exemplar. The calibration centre, clusters, and harmful/benign region selection still use the full corpus, so this is not a fully unseen test. We generate 256 tokens and ask Claude Opus 4.8 (medium reasoning effort) whether each harmful-prompt response delivers the requested content, such as instructions or code. A response counts as compliant if it supplies the requested content, even when accompanied by warnings, described as educational, or presented as protective advice. Refusals, content-free moralising, safe-topic deflections, and empty output do not count.

\begin{table}[htbp]
\centering
\small
\caption{Compliance after the region swap. Evaluated prompts are excluded from the harmful and benign region means used to calculate the swap, but remain in calibration, clustering, and region selection. Each rate is from 32 harmful prompts with 256-token generations judged by Claude Opus 4.8. L14 has the highest compliance among the four layers tested per model. At Llama L31, 30 of 32 generations are empty.}
\label{tab:region-intervention-compliance}
\begin{tabular}{lcccc}
\toprule
 & \multicolumn{2}{c}{Gemma-2-2B-it} & \multicolumn{2}{c}{Llama-3.1-8B-Instruct} \\
\cmidrule(lr){2-3}\cmidrule(lr){4-5}
comparison & layer & compliance & layer & compliance \\
\midrule
earlier        & L8  & 0.12 & L10 & 0.19 \\
L14            & L14 & \textbf{0.88} & L14 & \textbf{0.91} \\
best separated & L18 & 0.22 & L17 & 0.16 \\
deeper         & L24 & 0.16 & L31 & 0.03 \\
\bottomrule
\end{tabular}
\end{table}

The swap produces much more compliance at L14 than at the other sampled layers in both models. The layers that separate harmful and benign prompts most cleanly, Gemma L18 and Llama L17, score $0.22$ and $0.16$. Llama L31's $0.03$ reflects broken generation: 30 of 32 outputs are empty. We tested four layers per model, so we do not know the full range of layers at which the swap is effective. With 32 prompts and one model judge, borderline responses could also move the rates by a few prompts.

\paragraph{Controls.} We repeat the L14 intervention with three controls, generated in the same run as \Cref{tab:region-intervention-compliance} and judged the same way (\Cref{tab:region-intervention-controls}). Removing the harmful-region component without the benign add-back is directional ablation \citep{arditi2024refusal}. We also project off the span of as many benign region means as there are harmful regions, taking the first $n$ benign regions in region order (all $64$ benign regions for Llama L14, $29$ of $43$ for Gemma L14), and off as many random directions as there are harmful regions. The full swap results in much higher compliance than any of these controls. Projection alone induces compliance rates of $0.00$ in Gemma and $0.34$ in Llama. Projecting off the benign or random spans results in $0.00$ in both. The random span matches the number of directions, but not the amount removed from an activation, so it is only a limited control for a generic perturbation. Projection sets the activation's coordinate within the selected span to the calibration centre; the swap instead sets it to the benign regions' mean position there.

\begin{table}[htbp]
\centering
\small
\caption{Controls at L14, generated in the same run as \Cref{tab:region-intervention-compliance} with 256-token continuations and judged for compliance by Claude Opus 4.8. Region means exclude the evaluated prompts. The swap results in the highest compliance in both models; projection without the benign add-back creates some compliance in Llama but none in Gemma.}
\label{tab:region-intervention-controls}
\begin{tabular}{lcc}
\toprule
condition & Gemma-2-2B-it & Llama-3.1-8B-Instruct \\
\midrule
baseline (no intervention)             & 0.00 & 0.00 \\
project off harmful span (no add-back) & 0.00 & 0.34 \\
project off benign span                & 0.00 & 0.00 \\
random directions                      & 0.00 & 0.00 \\
full swap                              & \textbf{0.88} & \textbf{0.91} \\
\bottomrule
\end{tabular}
\end{table}

\section{Taboo Results}
\label{app:taboo}

\Cref{tab:taboo-full} shows one region for each secret. We choose the row after checking every region reached by the fine-tuned model but not by the base model at the sampled positions. Choosing the best row uses the known secret; building the candidate list does not. Token ranks start at zero.

\begin{table}[t]
\centering
\small
\caption{All 21 Taboo models at layer 32 and $p=12$. We chose each displayed region after looking for the secret in the candidate list. Member counts are fine-tuned/base activations; every displayed region has zero base members at the sampled positions.}
\label{tab:taboo-full}
\begin{tabular}{lrrrrr}
\toprule
Secret & $K$ & Region & Members & Embedding rank & Readout \\
\midrule
blue  & 61  & 48 & 10/0  & 0    & blue \\
book  & 87  & 11 & 60/0  & 18   & books, novels \\
cat   & 47  & 11 & 75/0  & 2    & feline, cats \\
chair & 42  & 10 & 25/0  & 3    & furniture, chairs \\
clock & 63  & 1  & 295/0 & 0    & clock, clocks \\
cloud & 49  & 11 & 5/0   & 1    & clouds, cloud \\
dance & 84  & 1  & 315/0 & 0    & dance, dancers \\
flag  & 61  & 53 & 10/0  & 0    & flag, flags \\
flame & 99  & 1  & 335/0 & 0    & flame, flames \\
gold  & 84  & 1  & 240/0 & 0    & gold \\
green & 74  & 1  & 315/0 & 0    & green \\
jump  & 63  & 1  & 275/0 & 1    & leap, jump \\
leaf  & 100 & 1  & 305/0 & 0    & leaf, foliage \\
moon  & 121 & 1  & 315/0 & 0    & moon, lunar \\
rock  & 58  & 27 & 5/0   & 562  & stadiums, CDs \\
salt  & 72  & 28 & 5/0   & 3,978 & unrelated \\
ship  & 76  & 1  & 295/0 & 0    & ship, ships \\
smile & 64  & 1  & 275/0 & 0    & smile \\
snow  & 72  & 1  & 255/0 & 0    & snow, snowfall \\
song  & 66  & 1  & 295/0 & 57   & musicians, singers \\
wave  & 63  & 1  & 285/0 & 1    & waves, wave \\
\bottomrule
\end{tabular}
\end{table}

Seventeen rows put the secret among their top four tokens; \emph{book} and \emph{song} give related words. Neither \emph{rock} nor \emph{salt} has a readable matching region at L32. We also checked L16, L20, L24, L28, L36, and L40 for those two models; their best secret ranks across these layers are 452 and 746. Neither secret is among the highest-ranked tokens at any of these layers.

Region occupancy alone does not reliably select the secret region. A rule that looks among the top two regions by excess fine-tuned occupancy for a top-four secret token succeeds in 13 of 21 models. It misses \emph{cat}, \emph{chair}, \emph{cloud}, and \emph{blue} despite readable regions elsewhere in their lists. 

We also checked sensitivity to resolution, layer, and the presence of the secret word in the input. On 60 transcripts each for \emph{gold}, \emph{chair}, and \emph{smile}, the region ranked first by excess fine-tuned occupancy still identifies the secret at both $p=4$ and $p=8$: the embedding ranks are $(0,2,2)$ at either resolution, and no base-model activation enters those regions. A layer sweep for \emph{gold} at $p=12$ results in ranks $2$, $20{,}623$, $8{,}144$, $1$, $0$, $0$, and $89{,}423$ at L16, L20, L24, L28, L32, L36, and L40. This result is strongest from L28 to L36; we have not swept every model. Finally, one of the 500 \emph{gold} transcripts contains the string \emph{gold}. Removing it and rebuilding still results in a rank-0 region with $280$ fine-tuned and no base members.

Both models are measured at the same two assistant-prefix positions, conditioned on the same user prompts and before the generated hints, so the candidate regions reflect a difference between the models on these sampled inputs. In a separate test with a model asked to guess the secret word (the auditor), adding decoded tokens from the top occupancy-ranked region helps on \emph{gold}, \emph{smile}, and \emph{moon}, but hurts on \emph{chair} and \emph{flag}. For \emph{chair}, the selected region reads as ``Sure''/``OK'' rather than the secret; for \emph{flag}, the auditor often guesses \emph{banner}. The benefit to the auditor therefore depends on which region is selected and how its readout is interpreted.

\section{Resolution-Separation Toy}
\label{app:resolution-toy}

At coarse resolution the dictionary groups prompts by one factor; at fine resolution it separates factors that occur together. We construct prompts of the form \texttt{``\{n\} (\{c1\}) \{c2\}''} over red, blue, and yellow, read the final-token activation from Gemma-2-2B L12, and calibrate the threshold on the toy itself. At coarse resolution, EP returns three regions, one for each prominent final colour (adjusted Rand index (ARI) $0.57$ against the ordered colour-pair labels). At finer resolution, it separates the ordered colour pairs: the sweep reaches ARI $0.99$ with seven regions, and a separate threshold search finds a six-region solution described below. When the final colour is restricted to red and blue, yellow anchors no region, but it still splits both prominent-colour regions at finer resolution.

We also use the toy to compare readouts. With three regions, the nearest region is always the prominent colour, the second-nearest is the background colour, and the third is the absent colour. A dense readout can therefore preserve overlap that a nearest-region readout discards, but the distances still do not form an additive decomposition. This result does not show that EP recovers long-range or abstract features; a factor that is always in the background may require a resolution so fine that the dictionary becomes difficult to read.

There are 300 prompts in the three-colour experiment, 50 for each ordered pair, and 200 in the experiment where yellow is never final. The integer ranges from 0 to 99, so prompts with the same colour pair still differ in their input text. The threshold is calibrated over the toy activations, and the build order is fixed.

\begin{table}[t]
\centering
\caption{Three-colour resolution sweep. Adjusted Rand index (ARI) measures agreement between regions and either ordered colour-pair labels or final-colour labels; $1$ indicates exact agreement.}
\begin{tabular}{rrrr}
\toprule
$p$ & $K$ & Colour-pair ARI & Final-colour ARI \\
\midrule
40 & 3 & 0.567 & 1.000 \\
22 & 5 & 0.821 & 0.725 \\
18 & 7 & 0.988 & 0.558 \\
16 & 9 & 0.952 & 0.529 \\
10 & 29 & 0.463 & 0.213 \\
\bottomrule
\end{tabular}
\end{table}

The separate threshold search finds a coarse setting, $p=33.5$, with three prominent-colour regions, and a fine setting, $p=21.5$, with six ordered-pair regions. For the region nearest the \texttt{(blue) red} condition, saved examples include \texttt{40 (blue) red} and \texttt{76 (yellow) red}; the latter belongs to the neighbouring condition, whose centroid is only $0.16$ away. In the experiment where yellow is never final, coarse $p=55$ constructs two regions and fine $p=35$ four: \texttt{(blue) red}, \texttt{(yellow) red}, \texttt{(red) blue}, and \texttt{(yellow) blue}.

\section{Future Work}
\label{app:future-work}

The following experiments would test where EP loses useful information and whether that information can be recovered.

\paragraph{Geometry and calibration.}
EP currently uses centred cosine distance and discards activation magnitude. Comparing it with Euclidean or covariance-weighted distance would show how much the results depend on the distance measure. The domain-saturation experiment could also be repeated with domain-specific calibration and longer stopping windows, to distinguish domain geometry from the effects of the threshold and stopping rule.

\paragraph{Labelling neighbouring regions.}
A region label often describes what its members share most strongly. Comparing neighbouring regions may reveal smaller differences that their individual labels miss. This could be tested by decoding differences between region representatives or asking an annotator to contrast their member examples.

\paragraph{Stability-filtered low-percentile dictionaries.}
We could build a fine dictionary and then filter its regions by stability (\S\ref{sec:stability}). A very small percentile ($p_1$, $p_2$, or smaller, down to $p_{0.5}$ or $p_{0.1}$) provides a large dictionary with many fine regions, including unstable ones. We would drop low-$D_i$ regions before evaluation. The possible benefit is that smaller percentiles give more regions to choose from, while $D_i$ filtering removes the noisiest among them. Whether this improves AxBench detection, sparse probing, top-$1$ probing, or SAEBench reconstruction is an empirical question; the current paper evaluates the full unfiltered dictionaries.

\paragraph{Neighbourhoods and hierarchy.}
The difference between the second-nearest and nearest exemplar distances measures how close an activation is to changing its assignment. It could be compared with classification errors, without assuming it is a calibrated confidence score. Grouping nearby exemplars would also give a hierarchy for inspection; we could then test whether the groups at each level have useful interpretations.

\section{Efficiency: Full Percentile Sweep}
\label{app:efficiency}

\Cref{tab:efficiency} reports per-percentile EP build budgets at both Gemma-2-2B settings used in the paper. Build-token counts exclude BOS and count the activations used for clustering. Each forward pass processes $128$ sequences of $128$ content tokens plus BOS; only the content-token activations enter the dictionary. The Gemma Scope baseline~\citep{lieberum2024gemmascope} consists of canonical 16k JumpReLU residual-stream SAEs on Gemma-2-2B, trained for $4 \times 10^{9}$ activation tokens at batch size $4{,}096$ ($\sim 10^{6}$ optimizer steps); total LLM forward passes for activation extraction and wall-clock training time are not reported in the source. The $p_1$ L20 EP build uses $3.62\times10^6$ construction tokens, about $1{,}100$ times fewer than this SAE training budget. Token counts are not a direct FLOP comparison: EP also performs exemplar-distance calculations, while the SAE optimises an encoder and decoder.

\begin{table}[htbp]
\caption{Per-percentile EP build budgets for the two settings used in the paper. EP performs no gradient descent at any stage, so no backward passes are needed.}
\label{tab:efficiency}
\centering
\small
\begin{tabular}{llrrr}
\toprule
Method & Layer & Dict.\ size & Build tokens & LLM fwd passes \\
\midrule
EP $p_{1}$  & L12 (base) & 20{,}295 & 15{,}089{,}664 & 921 \\
EP $p_{2}$  & L12 (base) & 5{,}129 & 5{,}029{,}888 & 307 \\
EP $p_{4}$  & L12 (base) & 1{,}193 & 1{,}015{,}808 & 62 \\
EP $p_{8}$  & L12 (base) & 325 & 475{,}136 & 29 \\
EP $p_{10}$ & L12 (base) & 203 & 344{,}064 & 21 \\
\midrule
EP $p_{0.5}$ & L20 (it) & 18{,}674 & 11{,}943{,}936 & 729 \\
EP $p_{1}$  & L20 (it) & 5{,}796 & 3{,}620{,}864 & 221 \\
EP $p_{2}$  & L20 (it) & 2{,}037 & 1{,}589{,}248 & 97 \\
EP $p_{4}$  & L20 (it) & 686 & 573{,}440 & 35 \\
EP $p_{8}$  & L20 (it) & 226 & 294{,}912 & 18 \\
\bottomrule
\end{tabular}
\end{table}

\section{Dictionary Builds}
\label{app:dictionaries}

Table~\ref{tab:efficiency} lists the saturated Pile dictionary builds used for AxBench and SAEBench in this paper. The task-specific dictionaries used for cross-model comparison, intervention, and Taboo are described in their respective sections. All except the Gemma-2-2B-it L20 $p_{0.5}$ build are \ifdefined\EPAnonymous available in the released dictionary collection.\else available in the Hugging Face dataset linked from the public repository.\fi All builds use sequences of $128$ tokens, batches of $128$ sequences, and random seed $0$. We extract an activation at every non-BOS token position and stop when a batch creates no new regions. The token counts therefore report the data consumed before stopping.

% NeurIPS-required checklist is part of the conference submission only,
% not the arXiv preprint. The filled-in checklist source lives with the
% NeurIPS submission, separate from this directory.

\end{document}